\documentclass[10pt, conference, compsocconf]{IEEEtran}
\usepackage{graphicx}
\usepackage{makecell}
\usepackage{multirow}
\usepackage{bbding}
\usepackage{xcolor}

\ifCLASSINFOpdf
\else
\fi
\usepackage[cmex10]{amsmath}
\usepackage{amsfonts}
\usepackage[font=footnotesize,caption=false]{subfig}
\graphicspath{ {./figures/} }

\begin{document}

\IEEEoverridecommandlockouts
\IEEEpubid{\makebox[\columnwidth]{19th Conference on Robots and Vision (CRV 2022), Toronto, Ontario, Canada. \hfill} \hspace{\columnsep}\makebox[\columnwidth]{ }}
%
\title{Monocular Robot Navigation with Self-Supervised Pretrained Vision Transformers}


\author{\IEEEauthorblockN{Miguel Saavedra-Ruiz$^*$}
\IEEEauthorblockA{DIRO, Mila - Quebec AI Institute\\
Université de Montréal\\
Montreal, Canada\\
miguel-angel.saavedra-ruiz@mila.quebec }
\and
\IEEEauthorblockN{Sacha Morin$^*$}
\IEEEauthorblockA{DIRO, Mila - Quebec AI Institute\\
Université de Montréal\\
Montreal, Canada\\
sacha.morin@mila.quebec }
\and
\IEEEauthorblockN{Liam Paull}
\IEEEauthorblockA{DIRO, Mila - Quebec AI Institute\\
Université de Montréal\\
Montreal, Canada\\
paulll@iro.umontreal.ca}
}


%


\maketitle

\begin{abstract}
In this work, we consider the problem of learning a perception model for monocular robot navigation using few annotated images. Using a Vision Transformer (ViT) pretrained with a label-free self-supervised method, we successfully train a coarse image segmentation model for the Duckietown environment using 70 training images. Our model performs coarse image segmentation at the 8x8 patch level, and the inference resolution can be adjusted to balance prediction granularity and real-time perception constraints.  We study how best to adapt a ViT to our task and environment, and find that some lightweight architectures can yield good single-image segmentation at a usable frame rate, even on CPU. The resulting perception model is used as the backbone for a simple yet robust visual servoing agent, which we deploy on a differential drive mobile robot to perform two tasks: lane following and obstacle avoidance.

\end{abstract}


%
\IEEEpeerreviewmaketitle

\section{Introduction}

In the past decade or so, deep learning has contributed to improving the state of the art on many computer vision tasks. Powered by architectures adapted to the image domain, like Convolutional Neural Networks (CNN) \cite{lecun1989backpropagation} or Vision Transformers (ViT) \cite{vit}, deep networks can successfully tackle tasks ranging from classification to dense semantic segmentation. While of clear interest to the design of vision-based robotics systems, deep learning still presents a number of limitations restricting its application to robotics. Firstly, deep learning methods can suffer from sample inefficiency and typically require a large number of annotated images to produce estimators with good generalization capabilities. Deployment in new visual environments are therefore likely to require an expensive data annotation procedure. Secondly, deep vision models have been characterized by a ``depth race'' with architectures of increasing size, which are of limited use for embodied agents requiring high inference performance on resource-constrained hardware.\textcolor{white}{\footnote{Equal contributions.}} 

In this work, we consider the problem of learning an instance segmentation model for monocular robot navigation using few annotated images. Our model performs ``coarse'' semantic segmentation by predicting labels at the \(8\times8\) patch level. We address the sample inefficiency issue by using standard data augmentation techniques as well as pretrained weights from a leading self-supervised method \cite{dino}. As for the computational aspect, we show how a ViT can be trained for the the task at hand and yield effective and efficient perceptual backbones for visual servoing. 

In summary, we make the following contributions: 
\begin{enumerate}
    \item Using a ViT pretrained with a label-free self-supervised method, we successfully train a coarse image segmentation model using only 70 training images;
    \item We show how the same model can be used to predict labels at different resolutions, allowing a compromise between prediction granularity, inference speed and memory footprint;
    \item We perform experimental validation by using the resulting segmentations for visual servoing of a real robot in the Duckietown environment.
\end{enumerate}
Our source code, annotated dataset and additional videos of our experiments can be found at \texttt{https://sachamorin.github.io/dino}. 

\section{Background}
\subsection{Vision Transformers}
ViTs~\cite{vit} have recently emerged as a competitor to CNNs~\cite{lecun1989backpropagation} for computer vision tasks~\cite{liu2022convnet}. Echoing the original transformer formulation for natural language processing \cite{transformer}, the ViT architecture decomposes images (``sentences'') into small image patches (``visual words''), typically of size \(8\times8\) or \(16\times16\). ViTs learn vector encodings for each patch via self-attention, that is, encodings are updated with a weighted combination of all other patch encodings. A single transformer layer can therefore learn dependencies between any two patches in the input image. 
For the specific task of image segmentation, transformer-based architectures usually rely on some encoder-decoder structure \cite{badrinarayanan2017segnet} where patch encodings are reconstructed or upsampled in some way to predict labels in pixel space \cite{segmenter, zheng2021rethinking, xie2021segformer, ranftl2021vision}.

\subsection{Self-Supervised Learning}

Self-Supervised Learning (SSL) is a subset of unsupervised learning which aims to learn generic and expressive data representations in the absence of labels. Recent successful applications of SSL include transformer-based language models like BERT or RoBERTa \cite{bert, roberta}. A similar trend exists for vision models, where SSL image representations achieve state-of-the-art performance on ImageNet benchmarks \cite{simclr, dino, maske-auto}. An important class of SSL algorithms for vision relies on contrastive learning, where the aim is to learn representations in which samples from the same class (``positive'') are close to one another and samples from distinct classes (``negative'') are separated. 

The traditional view of contrastive learning using positive and negative pairs was recently modified in BYOL~\cite{byol}, where the authors managed to train representations on the ImageNet dataset~\cite{imagenet} using only positive pairs and a teacher/student architecture.  
A followup to BYOL is DINO~\cite{dino}, which employs a similar architecture with a different loss. Interestingly, the authors show how self-supervised ViTs naturally learn explicit information about the semantic segmentation of an image in their attention masks, a finding that was not replicated in CNNs or in supervised ViTs.



Notwithstanding the undeniable success of SSL, there have been few attempts to extrapolate these ideas to real-world robots where data is abundant but labeling remains a time-consuming task. Some attempts have been made in the Reinforcement Learning (RL) community for navigation tasks~\cite{cc-rig, duckie-transformer}, but there has been little experimental validation of SSL representations for monocular image-based visual servoing \cite{servoing, thesis}. Additionally, standard evaluation procedures for SSL in computer vision tasks~\cite{swin, moco, color} 
do not encompass assessment in downstream applications for embodied agents in real-world environments. Our work differs from previous ones as we use pre-trained DINO weights in a limited data setting to navigate a robot. 

\subsection{Duckietown}

To assess the performance of our visual-servoing agent, we use the Duckietown platform \cite{duckietown}. Duckietown is an inexpensive open-source platform for autonomous driving education and research. Duckiebots (Fig. \ref{fig:mycroft}) are the main mobile agents and various tiles and accessories are available to design their driving environments. 
In this work we exploit the modularity of Duckietown to create two object-rich driving scenes composed of ducks, houses, signs and other Duckiebots (Fig. \ref{fig:lane_following} and \ref{fig:obstacles}). By modifying the layout of the scene after data collection, we can validate the robustness of the ViT segmentation predictions used for visual servoing.

\section{Coarse Semantic Segmentation with Vision Transformers}

\begin{figure}
    \centering
    \includegraphics[width=.49\textwidth]{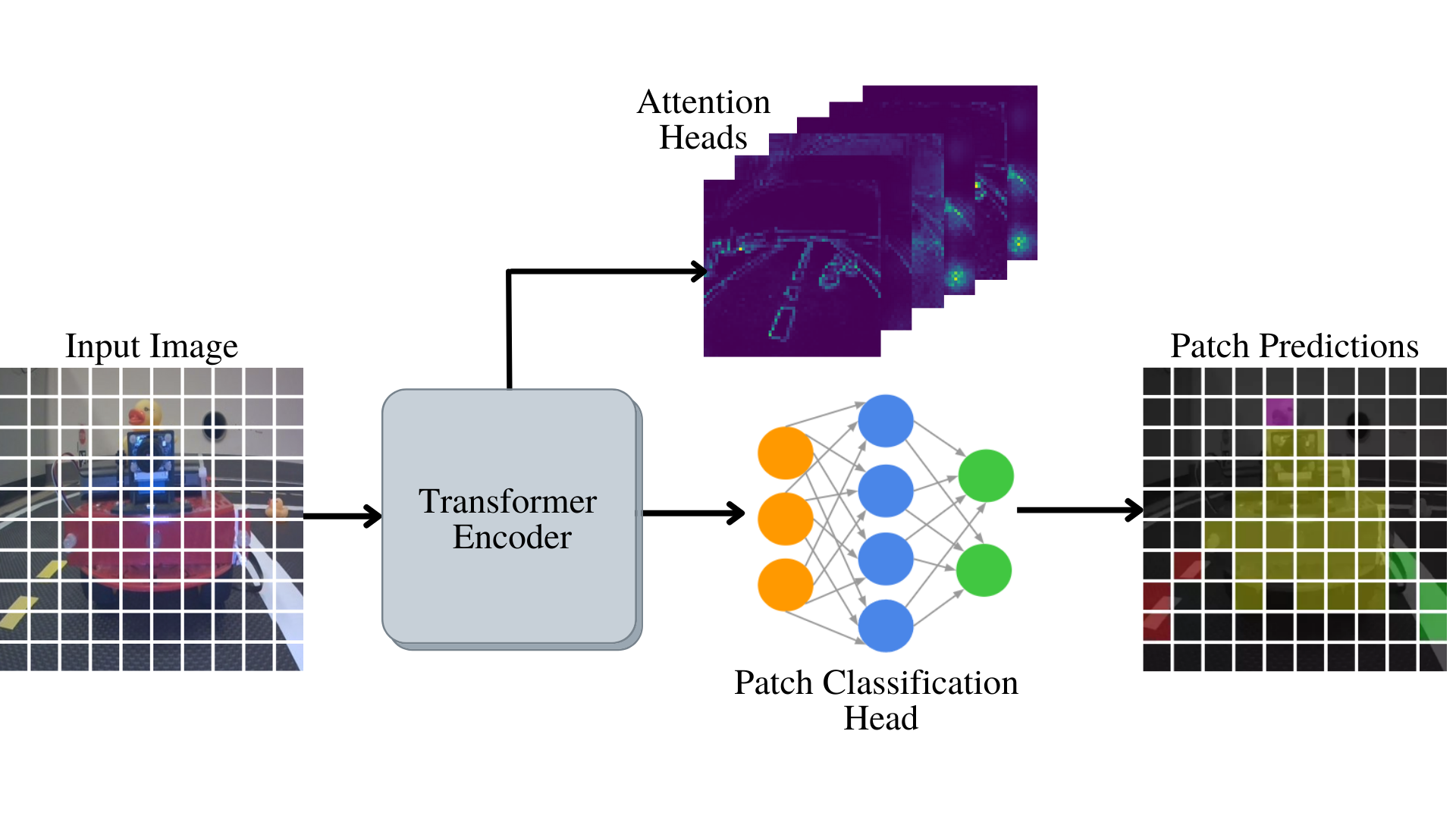}
    \caption{\textbf{Coarse semantic segmentation using ViTs.} We encode image patches using a ViT and predict a class label for each patch. We visualize attention heads to assess the effect of training in the transformer encoder.}
    \label{fig:pipeline}
\end{figure}

\begin{figure}
    \centering
    \includegraphics[width=.49\textwidth]{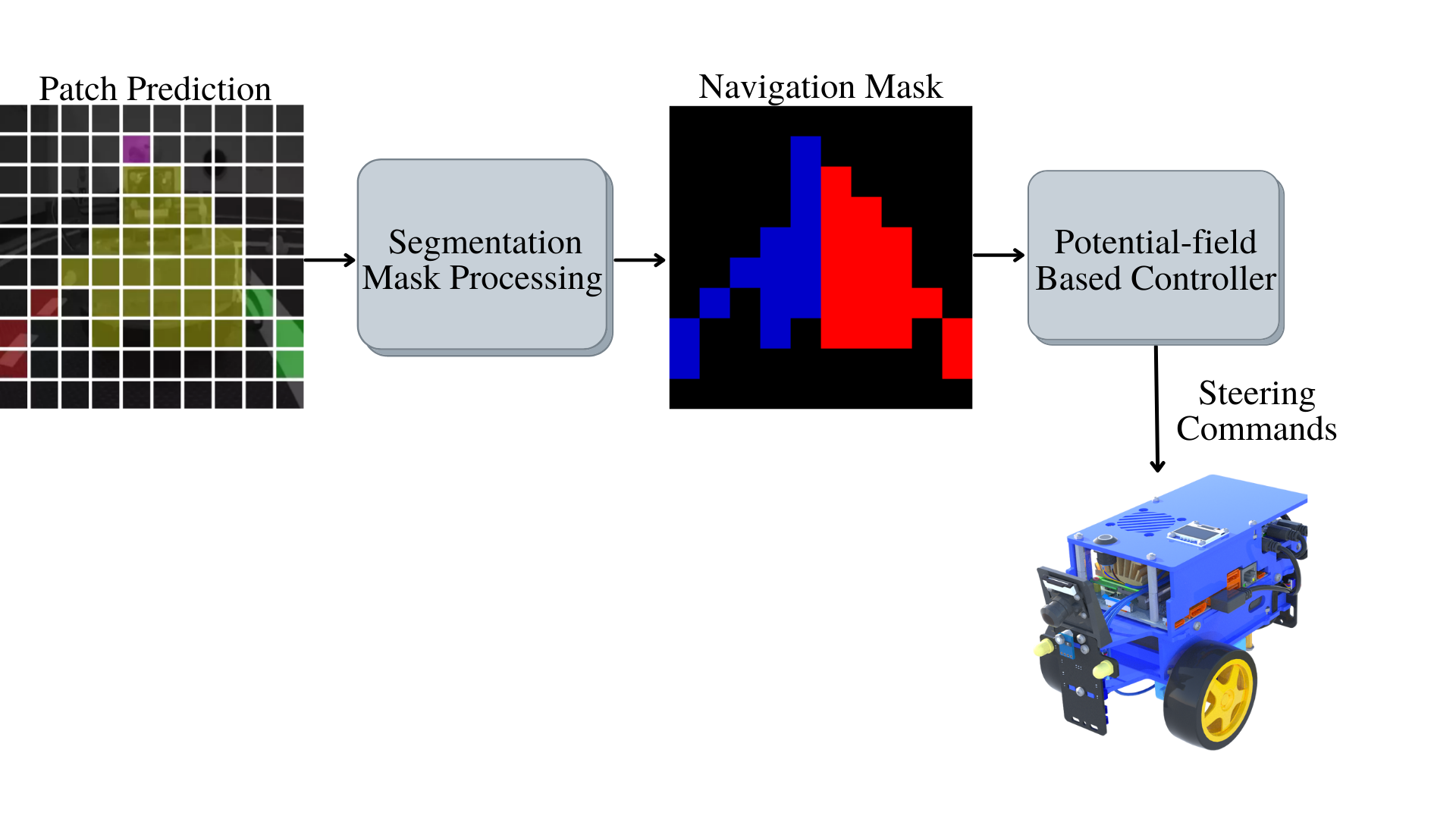}
    \caption{\textbf{Potential-Field based controller for lane following and obstacle avoidance.} The coarse segmentation output is used to compute a left (blue) and right (red) mask which are delivered to a potential-field based controller. The controller receives the mask and maps it as a ``repulsive'' potential to steer away from the half of the image with the most obstacle patches (Equation \ref{eq:loss}).}
    \label{fig:controller}
\end{figure}

Our approach is based on the following hypothesis: an agent can successfully and safely navigate an environment with low resolution segmentation masks. Thus, we propose to train a classifier to predict labels for every \(8 \times8 \) patch in an image. Our classifier is a fully-connected network which we apply over ViT patch encodings to predict a coarse segmentation mask (Fig. \ref{fig:pipeline}). For example, given a ViT with \(d\)-dimensional encodings, an 480p image will yield a \(60\times60\times d\) encoding map (\(480/8=60\)). During training, we therefore downsample the original ground truth mask accordingly using nearest neighbor interpolation to derive a single label for each patch. Our simple design avoids bespoke architectural components and allows to reuse pretrained weights from any vanilla ViT architecture, such as those from DINO~\cite{dino}.

\begin{table*}[!t]
\caption{Quantitative Assessment of Segmentation Quality and Speed}
\label{tab:inference}
\centering
\begin{tabular}{cccccccccc}
\hline
\makecell{Model} &Parameters  & \makecell{Input\\Resolution} &  \makecell{Output\\Resolution} & Patches &  \makecell{CPU \\Inference (Im./sec)} &  \makecell{GPU \\ Inference (Im./sec)} & \makecell{GPU\\RAM (GB)}  & mIoU & mAcc\\
\hline
\hline 
\multirow{3}{*}{ViT (1 block)} & \multirow{3}{*}{2.2M} & 240x240 & 30x30 & 900 & 58 & 168 & 0.10 &  0.71 & 0.79 \\
 &  & 480x480 & 60x60 & 3,600 & 6 & 47 & 0.75 & 0.78 & 0.85 \\
 &  & 960x960 & 120x120 & 14,400 & 0.5 & OOM & OOM & 0.75 & 0.83 \\
\hline
\multirow{3}{*}{ViT (3 block)} & \multirow{3}{*}{5.8M} & 240x240 & 30x30 & 900 & 23 & 131 & 0.11 & 0.76 & 0.84 \\
 &  & 480x480 & 60x60 & 3,600 & 2 & 29 & 0.93 & 0.86 & 0.90 \\
 &  & 960x960 & 120x120 & 14,400 & 0.2 & OOM & OOM & 0.84 & 0.89 \\
\hline
\multirow{3}{*}{CNN (24 layers)} & \multirow{3}{*}{1.6M} & 240x240 & 30x30 & 900 & 61 & 214 & 0.03 &  0.67 & 0.77 \\
 &  & 480x480 & 60x60 & 3,600 & 12 & 103 & 0.08 & 0.79 & 0.85 \\
 &  & 960x960 & 120x120 & 14,400 & 4 & 50 & 0.16 & 0.71 & 0.80 \\
\hline
\multirowcell{3}{CNN (32 layers)\\+ConvTranspose} & \multirow{3}{*}{7.1M} & 240x240 & 30x30 & 900 & 40 & 183 & 0.06 & 0.66 & 0.75 \\
 &  & 480x480 & 60x60 & 3,600 & 9 & 78 & 0.16 & 0.82 & 0.88 \\
 &  & 960x960 & 120x120 & 14,400 & 3 & 38 & 0.30 & 0.75 & 0.83 \\
\hline
\end{tabular}
\end{table*}

Our motivation for using ViTs is twofold: first, we hope to leverage the ability of transformers to learn long-range dependencies in an image using a relatively shallow architecture, which is an appealing proposition for real-time control tasks. Second, as with convolutional layers, a trained ViT can run segmentation on images at various resolution, as long as the patch size is the same. This implies that a ViT trained for coarse segmentation of \(8 \times 8\) patches will yield \(30 \times 30\) predictions for 240p images, \( 60 \times 60 \) predictions for 480p images, \(120 \times 120\) predictions for 960p images and so on. One can therefore adjust the granularity of the prediction and the associated computational load by simply downscaling or upscaling input images: a welcome flexibility for deployment on embodied agents with hardware constraints. ViTs can be trained and fine-tuned at different resolutions \cite{vit} and more advanced techniques for improving the performance at different test resolutions is a matter of active research \cite{xie2021segformer}.

\section{Image Segmentation Experiments}
\label{sec:segmentation}

\subsection{Dataset}
\begin{figure*}[!t]
\centerline{
\subfloat[\textbf{Robot.} Example of the differential drive DB21J Duckiebot model used in our experiments. While other sensors are available, we only used the camera for navigation. \label{fig:mycroft}]{\includegraphics[width=.45\textwidth,height=4.5cm,trim={0 22cm 0 30cm},clip]{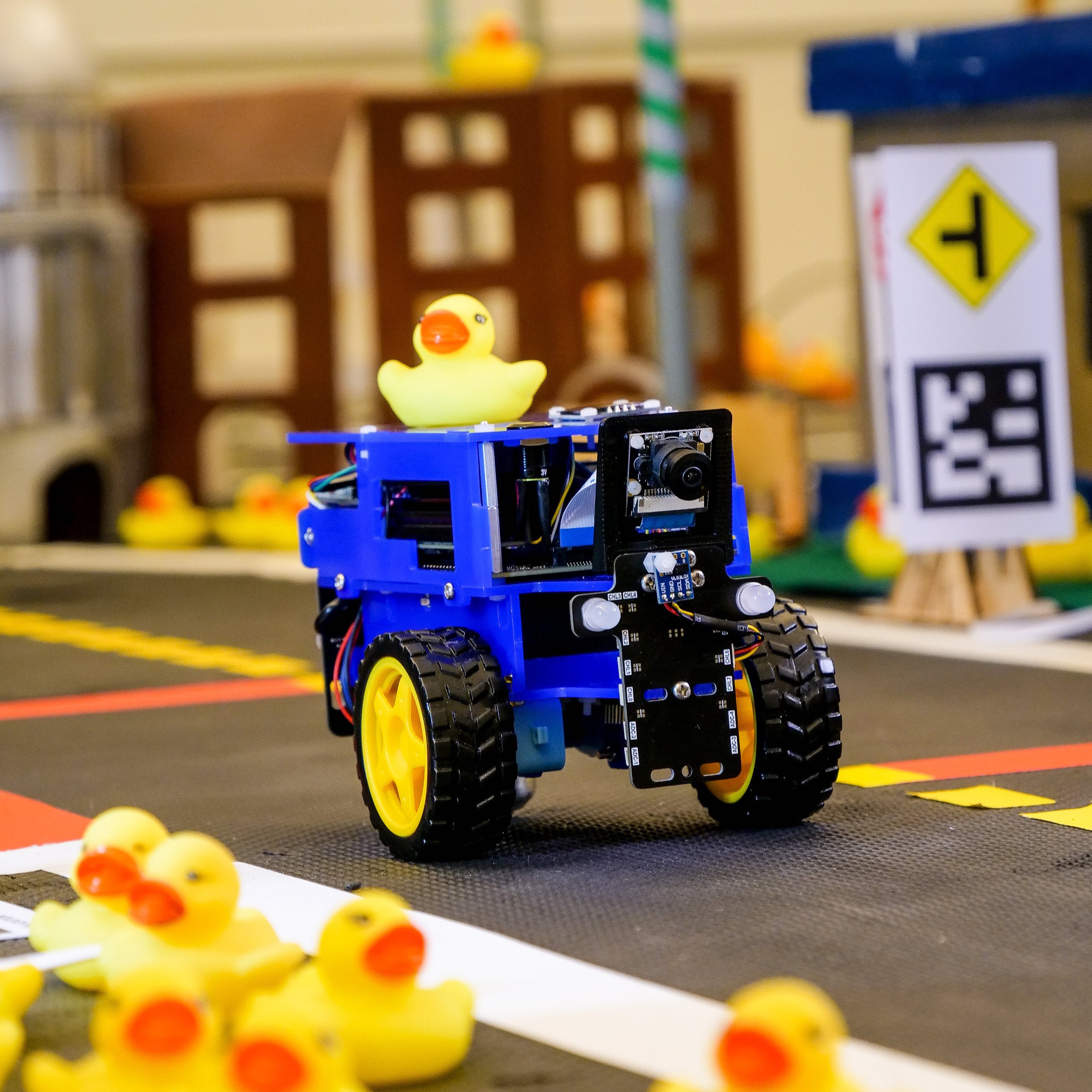}}
\hfil
\subfloat[\textbf{Data collection scene.} We collected 100 images by teleoperating the robot on the scene and occasionally adding various objects on the road. \label{fig:data_collection}]{\includegraphics[width=.45\textwidth,height=4.5cm]{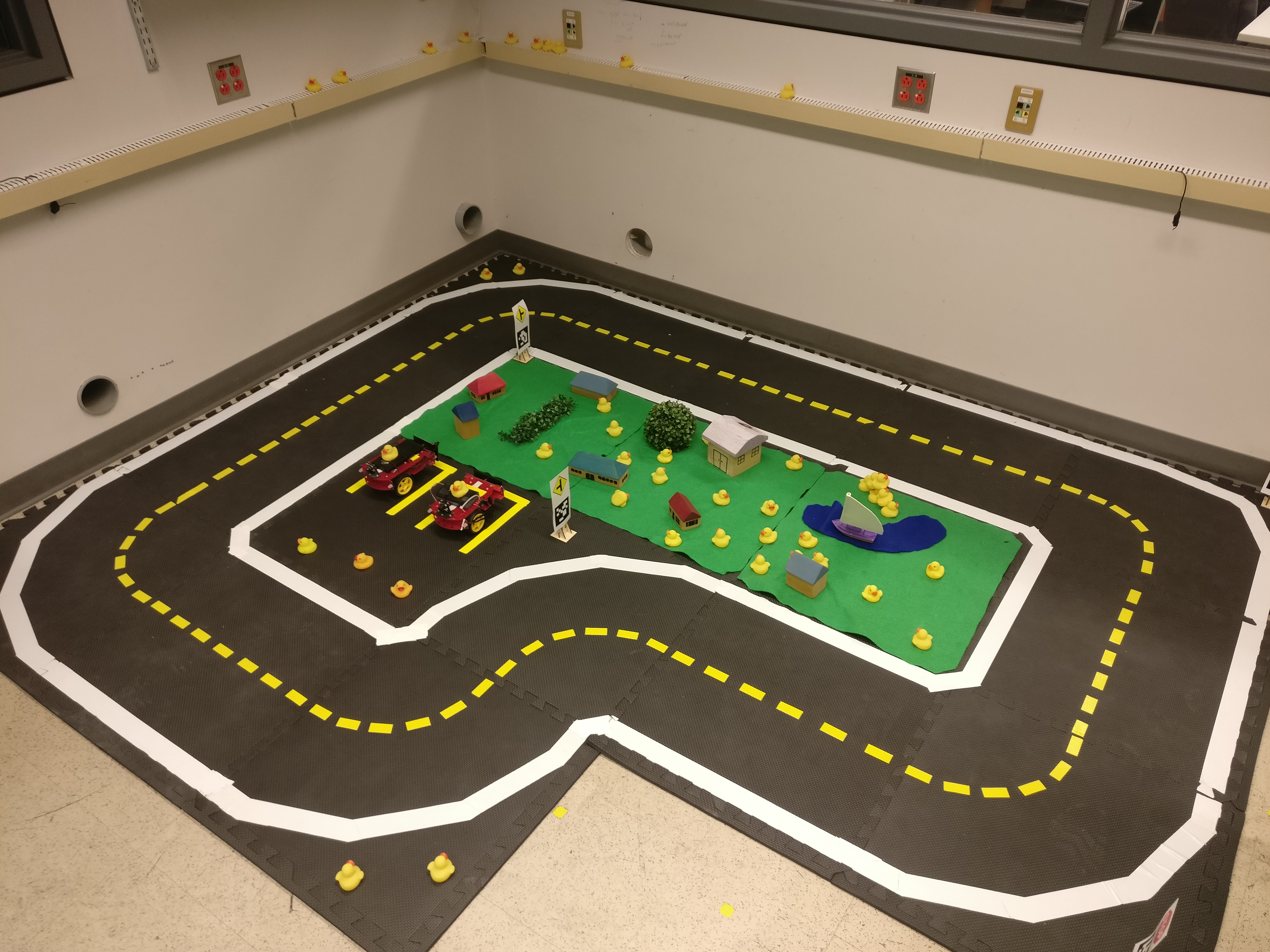}}
}
\centerline{
\subfloat[\textbf{Lane following scene.} The robot is tasked with completing laps without crossing the yellow line (left) or white line (right). We benchmarked both the outer and inner loops.  \label{fig:lane_following}]{\includegraphics[width=.45\textwidth,height=4.5cm]{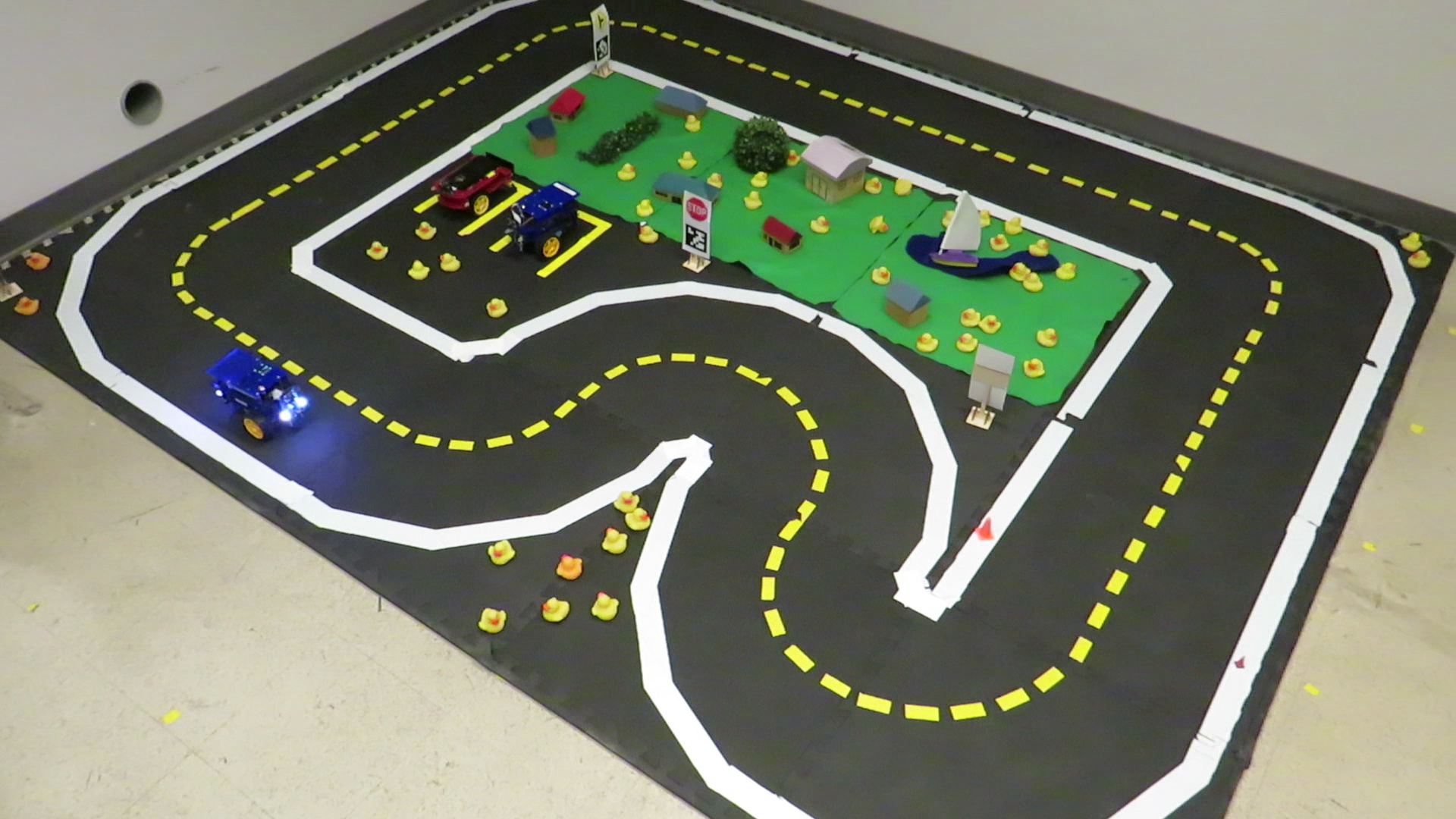}}
\hfil
\subfloat[\textbf{Obstacle avoidance scene.} The robot needs to complete laps without colliding with on-road obstacles. For this experiment, the controller ignores the yellow line predictions and the robot can navigate both lanes to avoid obstacles. In total, 4 duckiebots, 2 signs and 4 groups of duckies must be avoided during a loop. \label{fig:obstacles}]{\includegraphics[width=.45\textwidth,height=4.5cm]{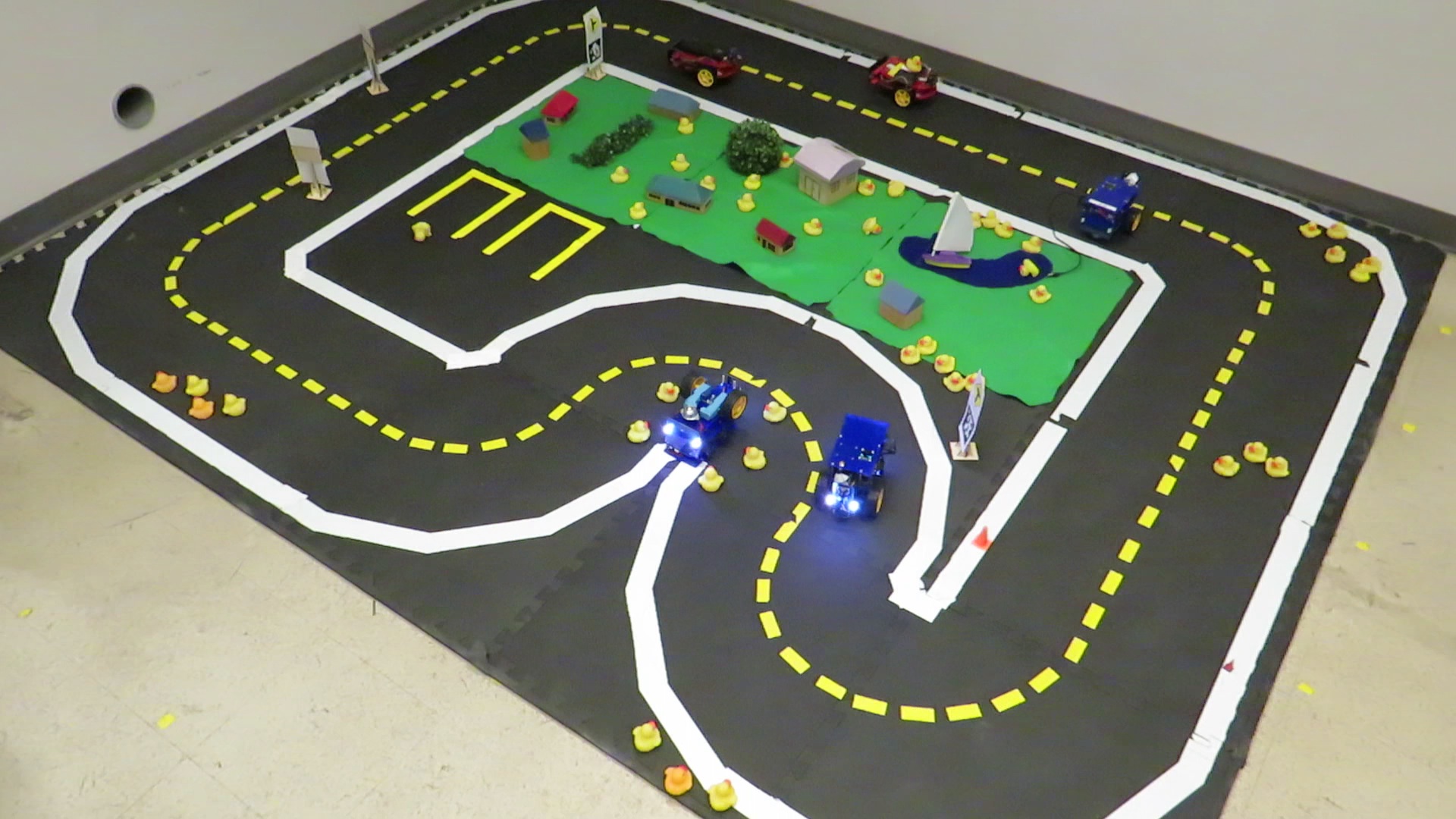}}
}
\caption{Experimental setup.}
\label{fig:data_scene}
\end{figure*}

Our dataset is composed of RGB images gathered using the on-board camera of our Duckiebot (Fig. \ref{fig:mycroft}). We collected images by teleoperating the robot in the data collection scene presented in Fig. \ref{fig:data_collection}. To increase the diversity in the training scene, various objects were added on the road during the episode. A total of 100 images were sampled from the recordings and labeled with 7 classes: Duckiebot, duckie, white lane marking, yellow lane marking, road sign and human hand. We use a 70-15-15 split for training, validation and testing.

\subsection{Model Training}
\begin{figure}
\centerline{
\includegraphics[width=.45\textwidth]{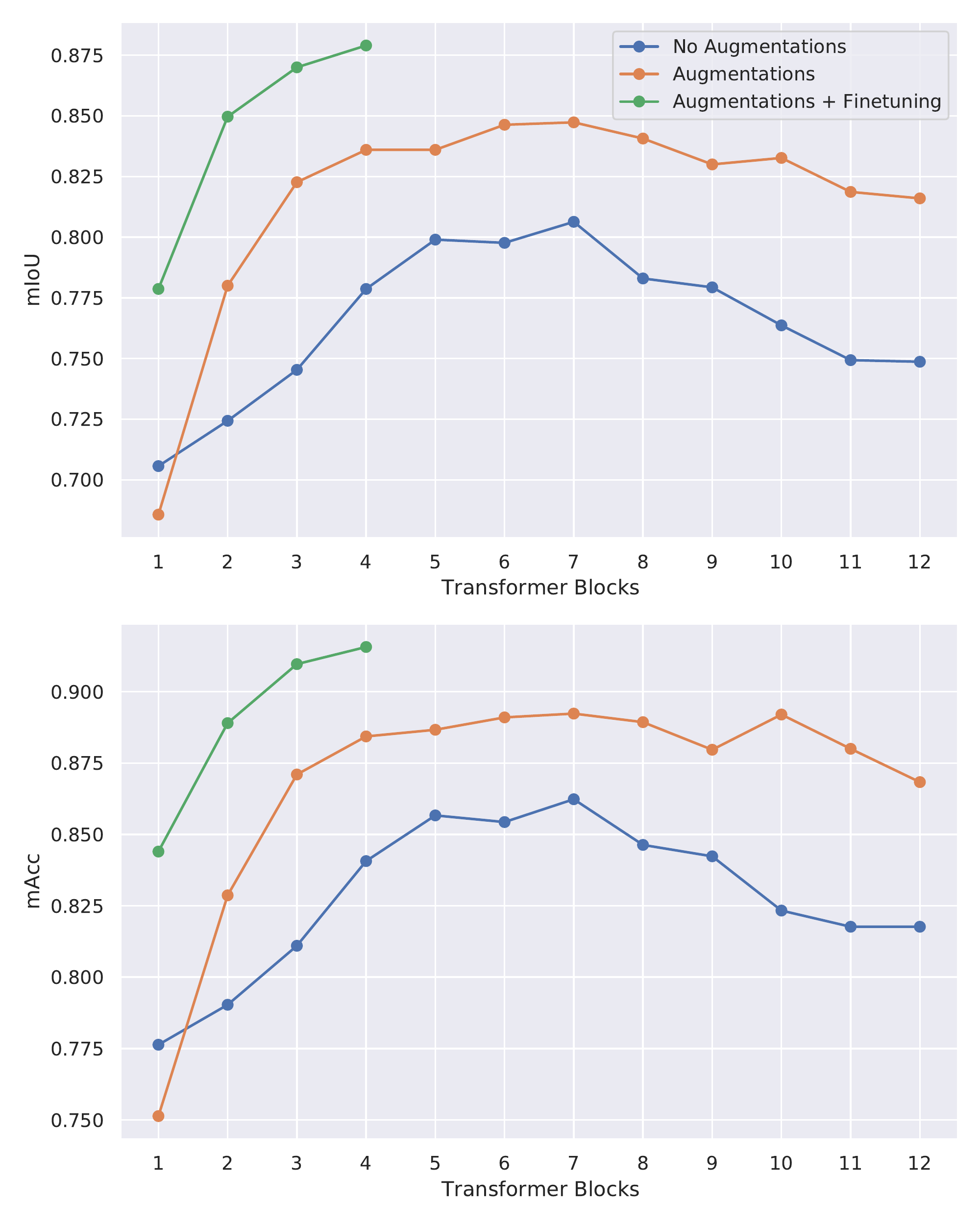}
}
\caption{Average class IoU (mIoU) and accuracy (mAcc) of various segmentation models on our Duckietown segmentation test dataset. Metrics are averaged over 3 seeds. The DINO architecture consists of 12 transformer blocks: we therefore probe the intermediary patch representations by training a segmentation head at various depths using a ViT backbone of $n$ blocks. While raw patch representations (\textbf{``No Augmentations''}) perform reasonably well, adding standard image augmentations (\textbf{``Augmentations''}) is hugely beneficial, despite the self-supervised DINO pretraining. In both the ``No Augmentations'' and ``Augmentations'' setup, the ViT backbone parameters are frozen and we only train the segmentation head. As expected, unfreezing the backbone and continuing training (\textbf{``Augmentations + Finetuning''}) increases both performance metrics and is particularly beneficial for the 1-block and 2-block backbones. We could not finetune backbones with more than 4 transformer blocks due to hardware constraints.}
\label{fig:blocks}
\end{figure}

We use the ViT architecture and pretrained weights from DINO for our perception backbone. Specifically, the ViT-S/8 architecture consists of 12 transformer blocks---a block includes a self-attention layer followed by 2 fully-connected layers---and takes \(8 \times 8\) patches as input. As with standard ViT architectures, the size of the feature map is constant throughout layers and any number of blocks will output predictions at the same resolution. Each patch is encoded as a 384-dimensional vector which we classify using a fully-connected segmentation head. In Fig. \ref{fig:blocks}, we vary the number of transformer blocks in the backbone as well as the augmentations applied to the training data in order to study segmentation performance. Encouragingly, we find that using only a few transformer blocks is sufficient to achieve good performance. We further observe that metrics plateau or even deteriorate with a backbone deeper than 5 blocks. Moreover, standard data augmentations and finetuning the backbone still appear necessary to maximize performance. The bigger part of the finetuning improvement is observed in the road sign class, which was not heavily featured in our dataset.

For training, we use 480p images, a batch size of 1 image (3600 \(8 \times 8\) patches), and train for 200 epochs of 1000 randomly sampled images. The validation set is used for checkpointing and the model with the best balanced validation accuracy is retained. When training the segmentation head only, we use the Adam optimizer with a learning rate of .001. For finetuning the backbone, we continue training for 200 epochs using the AdamW optimizer with a learning rate of 1e-4. We use standard data augmentations where mentioned: random crops, flips, shifts, scales, rotations, color jittering and Gaussian blur. Reported IoU and Accuracy scores are with respect to the downscaled ground truth masks, e.g., 480p predictions are benchmarked against the 60x60 interpolated masks to reflect patch-level performance.

\subsection{Inference}
In Table \ref{tab:inference}, we study the inference speed and quality of the finetuned 1-block and 3-block ViT segmentation models from the previous section. For comparison, we also study backbones built with the first layers of a DINO-pretrained ResNet-50 CNN architecture to perform the same task.  We benchmark the same models at different inference resolutions and find that the ViTs perform relatively well even on downscaled images. Importantly, all models can run inference at a reasonable framerate on GPU for 240p or 480p input. CPU inference at 240p would even be conceivable. All benchmarks were run on a \textit{11th Gen Intel Core i7-11800H @ 2.30GHz × 16} CPU and a \textit{NVIDIA GeForce RTX 3060} Laptop GPU. Contrary to the memory efficient CNNs, the ViT 960p resolution could not fit into GPU memory due to the unwieldy amount of patches.

\begin{figure*}[!t]
\includegraphics[trim={0 13cm 0 0},clip, width=1\textwidth]{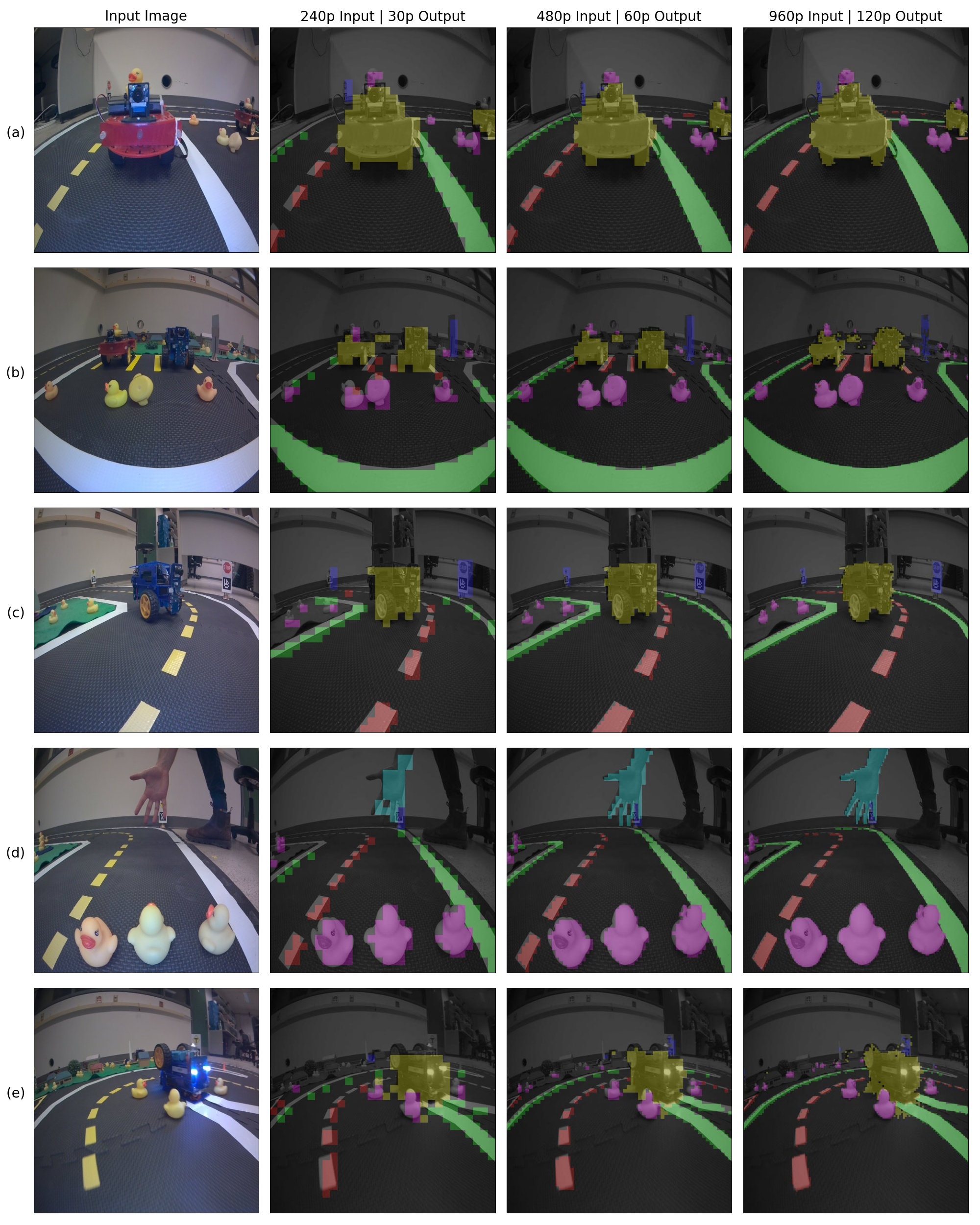}
\caption{Predictions of the \textbf{same 3-block ViT} at different resolutions. While the model was trained in the 480p regime, it performs well on downscaled (240p) or upscaled (960p) images. The 240p predictions are visually coarse, but accurate for nearby objects or large distant ones. We show in Subsections \ref{subsec:lane} and \ref{subsec:obstacles} how 240p and 480p predictions can be used for navigating a Duckietown environment. The 960p predictions are shown for illustrative purposes and are too slow for real-time navigation.}
\label{fig:predictions}
\end{figure*}

\begin{figure*}[!t] 
\centerline{
\begin{tabular}{ccc}
\includegraphics[width=.3\textwidth]{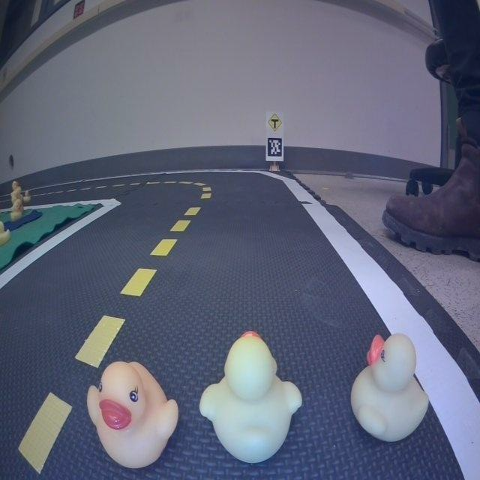} & \includegraphics[width=.3\textwidth]{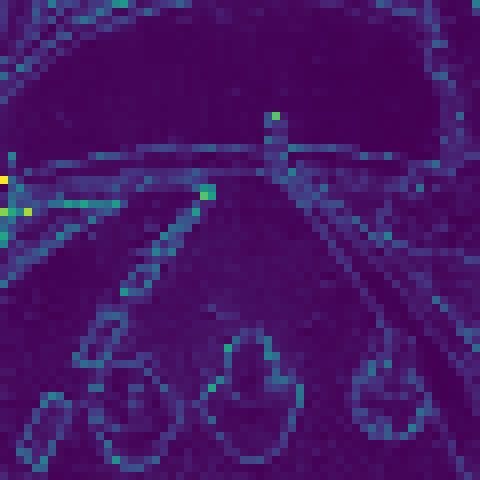} & \includegraphics[width=.3\textwidth]{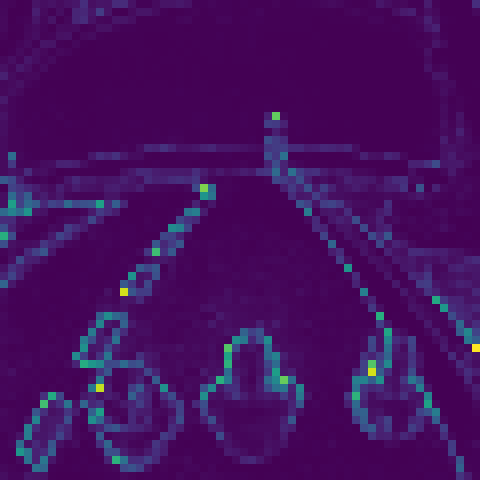}\\
\subfloat[Input image]{\includegraphics[width=.3\textwidth]{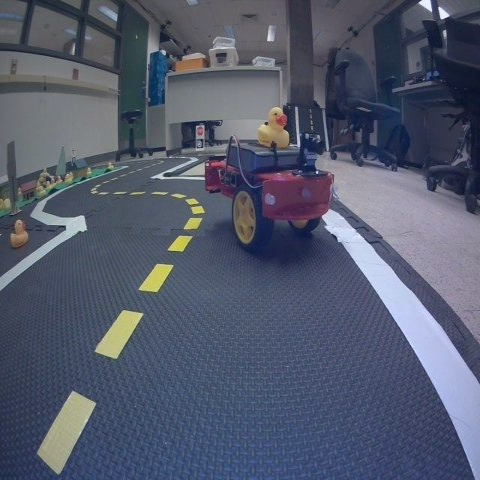}} & \subfloat[DINO]{\includegraphics[width=.3\textwidth]{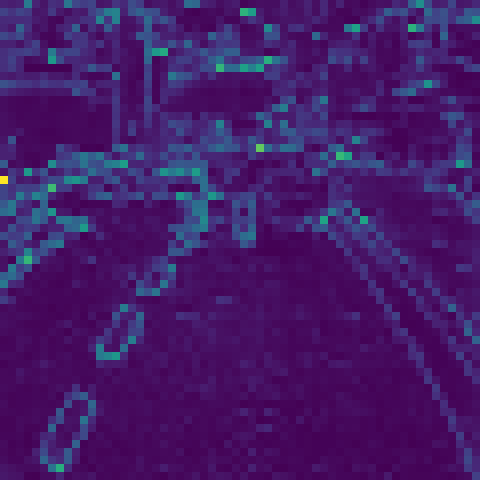}} & \subfloat[Finetuned]{\includegraphics[width=.3\textwidth]{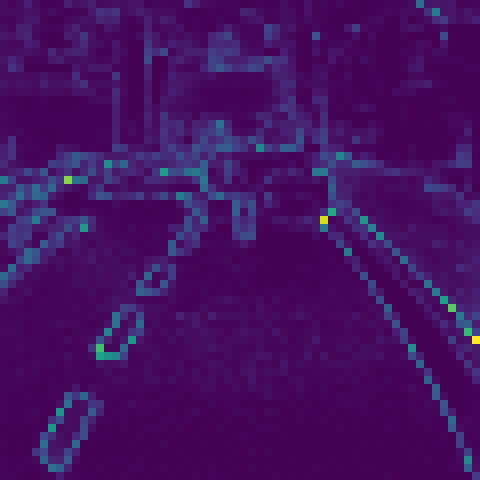}}
\end{tabular}
}
\caption{Segmentation masks of the DINO 3-block pretrained model comapared with a finetuned version. Results demonstrate how the attention masks of the pretrained model attend to a general variety of objects in the scene like the desk in row two. Interestingly, the attention masks of the finetuned model attend primarily to the objects that belong to one of the classes in our dataset, ignoring objects that are not relevant to our task.}
\label{fig:attention}
\end{figure*}

ViT predictions can be further visualized in Fig. \ref{fig:predictions}. Additionally, we compare the attention masks of the pretrained 3-block versus the finetuned one in Fig. \ref{fig:attention} where interestingly, the finetuned model learns to attend mostly to those objects that are relevant to our task.

\section{Navigation Experiments}
\label{sec:navigation}

In this section, we assess the performance of the trained 1-block and 3-block ViT models from Table \ref{tab:inference} on two different monocular visual servoing tasks: lane following and obstacles avoidance. The robot is controlled using a potential-field based controllers which receives as input the segmentation output produced by a trained ViT and outputs steering commands for the robot while maintaining a fixed linear velocity as shown in Fig. \ref{fig:controller}. For the lane-following task the goal is to maintain the vehicle centered in a lane while in the obstacle avoidance scene, the agent can use both lanes to navigate and avoid obstacles. 

The scenes for both experiments are modified from the original data collection scene. In Fig. \ref{fig:lane_following}, we show the driving environment used for lane-following where an additional U-turn is added to measure the robustness of the driving agent. The same driving scene is reused for the obstacle avoidance problem but different objects are added on the road (Fig. \ref{fig:obstacles}).

\subsection{Lane following}
\label{subsec:lane}


The objective of this task is to navigate the vehicle on the road (without obstacles) and maintain the vehicle centered between the yellow and white lines. We first obtain segmentation predictions \(\boldsymbol{S}_t \in \mathbb{R}^{n \times n}\) for the current $n\times n$ resized camera frame, then compute a navigation mask \(\boldsymbol{M}_t \in \{0,1\}^{n \times n}\) by extracting the pixel patches corresponding to white and yellow lines. This mask is used to compute a ``repulsive'' potential pushing the agent to steer away from the half of the image with the most line patches. More formally, the steering angle is controlled based on



\begin{equation}
\label{eq:loss}
    \phi_{t+1} = \phi_{t} - \gamma \sum_{i, j} (\boldsymbol{P} \odot \boldsymbol{M}_t)_{i, j}.
\end{equation}

where $P$ is a sign mask with $-1$ values on the left and $+1$ values on the right, and $\gamma$ is a weighting factor. This purely reactive controller will reach an equilibrium by keeping the same energy on the right and the left, i.e., by being centered between the white and yellow lines. We present a visualization of the controller in Fig. \ref{fig:controller}. 

Each model is evaluated by navigating the agent for five loops (2 outer loops and 3 inner loops) and report the number of minor and major infractions. A minor infraction occurs when the robot steps over either the white or yellow line. A major infraction is defined as any event requiring human intervention to put the agent back on track. Both models are evaluated at 240p and 480p input resolutions and compared against the standard lane following system implemented in Duckietown, which consists of HSV filters alongside a particle filter for state estimation.

The results of the lane following evaluation are reported in Table \ref{table-following}. Both models perform equally well in the outer loop, however, 
the inner loop proves more challenging.
The best performing model was with 1-block at 240p with a total of four minor infractions and zero major infractions. We hypothesize this good performance is owed to the high throughput of the model, which allows for better reaction time in the controller. Surprisingly, the high-capacity 3-block model 
reports a higher number of infractions. The baseline model is the worst-performing 
whose result is a likely consequence of the HSV filters producing false positives for line detection when white or yellow objects are placed in the scene (see Fig. \ref{fig:lane_following}).

\begin{table}[ht]
\caption{Lane following results}
\label{table-following}
\centering
\begin{tabular}{cccccc}
\hline
\makecell{Model} & \makecell{Input\\Resolution} & Outer & Inner  & \makecell{Minor\\Infractions} & \makecell{Major\\Infractions} \\
\hline
\hline 
\multirow{2}{*}{Baseline} & \multirow{2}{*}{480p} & \Checkmark &  & 7 & 0 \\
 &  &  & \Checkmark & 11 & 2 \\
 \hline
\multirow{4}{*}{\makecell{ViT\\1 block}} & \multirow{2}{*}{240p} & \Checkmark &  & 1 & 0 \\
 &  &  & \Checkmark & 3 & 0 \\
 & \multirow{2}{*}{480p}  & \Checkmark & & 1 & 0 \\
 &  &  & \Checkmark & 4 & 1 \\
\hline
\multirow{4}{*}{\makecell{ViT\\3 blocks}} & \multirow{2}{*}{240p} & \Checkmark &  & 1 & 0 \\
 &  &  & \Checkmark & 4 & 3 \\
 & \multirow{2}{*}{480p} & \Checkmark & & 1 & 0 \\
 &  &  & \Checkmark & 8 & 0 \\
\hline
\end{tabular}
\end{table}

These results demonstrate that even though higher capacity models perform better than low capacity ones with respect to test segmentation metrics, more variables should be considered for real-world deployment. Additionally, while the coarse segmentation predictions (see Fig. \ref{fig:predictions} column~2) lower the resolution to increase inference speed, they still appear to hold enough information to accurately navigate the agent within the environment. 


\subsection{Obstacle Avoidance}

\label{subsec:obstacles}

The objective for the obstacles avoidance experiment is to navigate the agent through the scene in Fig. \ref{fig:obstacles} while avoiding on-road duckies, signs and Duckiebots. This task was designed to validate the performance of the agent within a more challenging environment, and make use of predictions for all classes. We define the area  between both white lines as being drivable, i.e., the agent is allowed to cross the yellow line to avoid obstacles without penalty. Particularly, the controller is the same used in Subsection \ref{subsec:lane} with the difference that the navigation mask now includes all obstacles to be avoided. Therefore, pixels classified as white line, ducks, signs and duckiebots are actively contributing to the ``repulsive'' potential. This potential function encourages the agent to drive on the road while avoiding objects placed on it. The assessment here is similar to that of Subsection \ref{subsec:lane} (same number of loops) with the difference that driving over the yellow line is permitted and small contacts with obstacles are considered minor infractions.

\begin{table}[ht]
\caption{Obstacles avoidance results}
\label{table-obstacles}
\centering
\begin{tabular}{cccccc}
\hline
\makecell{ViT\\Blocks} & \makecell{Input\\Resolution}  & \makecell{Minor\\Infractions} & \makecell{Major\\Infractions} \\
\hline
\hline 
\multirow{2}{*}{1} & 240p & 2 & 0 \\
 & 480p & 1 & 2 \\
\hline
\multirow{2}{*}{3} & 240p & 1 & 4 \\
 & 480p & 3 & 1 \\
\hline
\end{tabular}
\end{table}

We evaluate the same models as in the previous subsection (1-block, 3-block, at 240p and 480p). For this benchmark, we do not have a particular baseline in the Duckietown stack to compare with. The results are presented in Table \ref{table-obstacles} and are consistent with the ones in Table \ref{table-following}. The best performing model was once again 1-block with an input resolution of 240p, reporting a total of two minor infractions and zero major ones. The high-capacity models did not perform as-well as expected even though the segmentation results produced by those are of superior quality, again suggesting that visual servoing benefits from the higher framerate of the shallow 1-block backbone.

\section{Conclusion}

In this work, we study how embodied agents with vision-based motion can benefit from ViTs pretrained via SSL methods. Specifically, we train a perception model with only 70 images to navigate a real robot in two monocular visual-servoing tasks. Additionally, in contrast to previous SSL literature for general computer vision tasks, our agent appears to benefit more from small high-throughput models rather than large high-capacity ones. We demonstrate how ViT architectures can flexibly adapt their inference resolution based on available resources, and how they can be used in robotic application depending on the precision needed by the embodied agent. Our approach is based on predicting labels for 8x8 image patches, and is not well-suited for predicting high-resolution segmentation masks, in which case an encoder-decoder architecture should be preferred. The low resolution of our predictions does not seem to hinder navigation performance however, and we foresee as an interesting research direction how those high-throughput low-resolution predictions affect safety-critical applications by scaling our method to more challenging scenarios. Moreover, training perception  models in an SSL fashion on sensory data from the robot itself rather than generic image datasets (e.g., ImageNet) appears to be a promising research avenue, and is likely to yield visual representations that are better adapted to downstream visual servoing applications.

\section*{Acknowledgment}
The authors would like to thank Gustavo Salazar and Lilibeth Escobar for their help labeling the dataset. Special thanks to Charlie Gauthier for her help setting-up the Duckietown experiments. This research was partially funded by an IVADO (l'Institut de valorisation des donn\'{e}es) MSc. Scholarship and an FRQNT (Fonds de recherche du Québec – Nature et technologies) B1X Scholarship [\emph{S.M.}].



\bibliographystyle{IEEEtran}
\bibliography{ref}

\begin{thebibliography}{10}
\providecommand{\url}[1]{#1}
\csname url@samestyle\endcsname
\providecommand{\newblock}{\relax}
\providecommand{\bibinfo}[2]{#2}
\providecommand{\BIBentrySTDinterwordspacing}{\spaceskip=0pt\relax}
\providecommand{\BIBentryALTinterwordstretchfactor}{4}
\providecommand{\BIBentryALTinterwordspacing}{\spaceskip=\fontdimen2\font plus
\BIBentryALTinterwordstretchfactor\fontdimen3\font minus
  \fontdimen4\font\relax}
\providecommand{\BIBforeignlanguage}[2]{{%
\expandafter\ifx\csname l@#1\endcsname\relax
\typeout{** WARNING: IEEEtran.bst: No hyphenation pattern has been}%
\typeout{** loaded for the language `#1'. Using the pattern for}%
\typeout{** the default language instead.}%
\else
\language=\csname l@#1\endcsname
\fi
#2}}
\providecommand{\BIBdecl}{\relax}
\BIBdecl

\bibitem{lecun1989backpropagation}
Y.~LeCun, B.~Boser, J.~S. Denker, D.~Henderson, R.~E. Howard, W.~Hubbard, and
  L.~D. Jackel, ``Backpropagation applied to handwritten zip code
  recognition,'' \emph{Neural computation}, vol.~1, no.~4, pp. 541--551, 1989.

\bibitem{vit}
A.~Dosovitskiy, L.~Beyer, A.~Kolesnikov, D.~Weissenborn, X.~Zhai,
  T.~Unterthiner, M.~Dehghani, M.~Minderer, G.~Heigold, S.~Gelly \emph{et~al.},
  ``An image is worth 16x16 words: Transformers for image recognition at
  scale,'' \emph{arXiv preprint arXiv:2010.11929}, 2020.

\bibitem{dino}
M.~Caron, H.~Touvron, I.~Misra, H.~Jégou, J.~Mairal, P.~Bojanowski, and
  A.~Joulin, ``Emerging properties in self-supervised vision transformers,''
  2021.

\bibitem{liu2022convnet}
Z.~Liu, H.~Mao, C.-Y. Wu, C.~Feichtenhofer, T.~Darrell, and S.~Xie, ``A convnet
  for the 2020s,'' \emph{arXiv preprint arXiv:2201.03545}, 2022.

\bibitem{transformer}
A.~Vaswani, N.~Shazeer, N.~Parmar, J.~Uszkoreit, L.~Jones, A.~N. Gomez,
  {\L}.~Kaiser, and I.~Polosukhin, ``Attention is all you need,'' in
  \emph{Advances in neural information processing systems}, 2017, pp.
  5998--6008.

\bibitem{badrinarayanan2017segnet}
V.~Badrinarayanan, A.~Kendall, and R.~Cipolla, ``Segnet: A deep convolutional
  encoder-decoder architecture for image segmentation,'' \emph{IEEE
  transactions on pattern analysis and machine intelligence}, vol.~39, no.~12,
  pp. 2481--2495, 2017.

\bibitem{segmenter}
R.~Strudel, R.~Garcia, I.~Laptev, and C.~Schmid, ``Segmenter: Transformer for
  semantic segmentation,'' in \emph{Proceedings of the IEEE/CVF International
  Conference on Computer Vision}, 2021, pp. 7262--7272.

\bibitem{zheng2021rethinking}
S.~Zheng, J.~Lu, H.~Zhao, X.~Zhu, Z.~Luo, Y.~Wang, Y.~Fu, J.~Feng, T.~Xiang,
  P.~H. Torr \emph{et~al.}, ``Rethinking semantic segmentation from a
  sequence-to-sequence perspective with transformers,'' in \emph{Proceedings of
  the IEEE/CVF conference on computer vision and pattern recognition}, 2021,
  pp. 6881--6890.

\bibitem{xie2021segformer}
E.~Xie, W.~Wang, Z.~Yu, A.~Anandkumar, J.~M. Alvarez, and P.~Luo, ``Segformer:
  Simple and efficient design for semantic segmentation with transformers,''
  \emph{Advances in Neural Information Processing Systems}, vol.~34, 2021.

\bibitem{ranftl2021vision}
R.~Ranftl, A.~Bochkovskiy, and V.~Koltun, ``Vision transformers for dense
  prediction,'' in \emph{Proceedings of the IEEE/CVF International Conference
  on Computer Vision}, 2021, pp. 12\,179--12\,188.

\bibitem{bert}
J.~Devlin, M.-W. Chang, K.~Lee, and K.~Toutanova, ``Bert: Pre-training of deep
  bidirectional transformers for language understanding,'' in \emph{NAACL},
  2019.

\bibitem{roberta}
Y.~Liu, M.~Ott, N.~Goyal, J.~Du, M.~Joshi, D.~Chen, O.~Levy, M.~Lewis,
  L.~Zettlemoyer, and V.~Stoyanov, ``Roberta: A robustly optimized bert
  pretraining approach,'' \emph{ArXiv}, vol. abs/1907.11692, 2019.

\bibitem{simclr}
T.~Chen, S.~Kornblith, M.~Norouzi, and G.~Hinton, ``A simple framework for
  contrastive learning of visual representations,'' Feb. 2020.

\bibitem{maske-auto}
K.~He, X.~Chen, S.~Xie, Y.~Li, P.~Doll{\'a}r, and R.~Girshick, ``Masked
  autoencoders are scalable vision learners,'' \emph{arXiv preprint
  arXiv:2111.06377}, 2021.

\bibitem{byol}
J.-B. Grill, F.~Strub, F.~Altch\'{e}, C.~Tallec, P.~Richemond, E.~Buchatskaya,
  C.~Doersch, B.~Avila~Pires, Z.~Guo, M.~Gheshlaghi~Azar, B.~Piot,
  k.~kavukcuoglu, R.~Munos, and M.~Valko, ``Bootstrap your own latent - a new
  approach to self-supervised learning,'' in \emph{Advances in Neural
  Information Processing Systems}, H.~Larochelle, M.~Ranzato, R.~Hadsell, M.~F.
  Balcan, and H.~Lin, Eds., vol.~33.\hskip 1em plus 0.5em minus 0.4em\relax
  Curran Associates, Inc., 2020, pp. 21\,271--21\,284.

\bibitem{imagenet}
O.~Russakovsky, J.~Deng, H.~Su, J.~Krause, S.~Satheesh, S.~Ma, Z.~Huang,
  A.~Karpathy, A.~Khosla, M.~Bernstein \emph{et~al.}, ``Imagenet large scale
  visual recognition challenge,'' \emph{International journal of computer
  vision}, vol. 115, no.~3, pp. 211--252, 2015.

\bibitem{cc-rig}
A.~Nair, S.~Bahl, A.~Khazatsky, V.~Pong, G.~Berseth, and S.~Levine,
  ``Contextual imagined goals for self-supervised robotic learning,'' in
  \emph{Proceedings of the Conference on Robot Learning}, ser. Proceedings of
  Machine Learning Research, L.~P. Kaelbling, D.~Kragic, and K.~Sugiura, Eds.,
  vol. 100.\hskip 1em plus 0.5em minus 0.4em\relax PMLR, 30 Oct--01 Nov 2020,
  pp. 530--539.

\bibitem{duckie-transformer}
W.~Shi, G.~Huang, S.~Song, Z.~Wang, T.~Lin, and C.~Wu, ``Self-supervised
  discovering of interpretable features for reinforcement learning,''
  \emph{IEEE Transactions on Pattern Analysis and Machine Intelligence}, 2020.

\bibitem{servoing}
J.~Dong and J.~Zhang, ``A new image-based visual servoing method with velocity
  direction control,'' \emph{Journal of the Franklin Institute}, vol. 357,
  no.~7, pp. 3993--4007, 2020.

\bibitem{thesis}
M.~Saavedra-Ruiz, A.~M. Pinto-Vargas, and V.~Romero-Cano, ``Monocular visual
  autonomous landing system for quadcopter drones using software in the loop,''
  \emph{IEEE Aerospace and Electronic Systems Magazine}, pp. 1--1, 2021.

\bibitem{swin}
Z.~Liu, Y.~Lin, Y.~Cao, H.~Hu, Y.~Wei, Z.~Zhang, S.~Lin, and B.~Guo, ``Swin
  transformer: Hierarchical vision transformer using shifted windows,'' in
  \emph{Proceedings of the IEEE/CVF International Conference on Computer
  Vision}, 2021, pp. 10\,012--10\,022.

\bibitem{moco}
K.~He, H.~Fan, Y.~Wu, S.~Xie, and R.~Girshick, ``Momentum contrast for
  unsupervised visual representation learning,'' in \emph{Proceedings of the
  IEEE/CVF conference on computer vision and pattern recognition}, 2020, pp.
  9729--9738.

\bibitem{color}
R.~Zhang, P.~Isola, and A.~A. Efros, ``Colorful image colorization,'' in
  \emph{European conference on computer vision}.\hskip 1em plus 0.5em minus
  0.4em\relax Springer, 2016, pp. 649--666.

\bibitem{duckietown}
L.~Paull, J.~Tani, H.~Ahn, J.~Alonso-Mora, L.~Carlone, M.~Cap, Y.~F. Chen,
  C.~Choi, J.~Dusek, Y.~Fang \emph{et~al.}, ``Duckietown: an open, inexpensive
  and flexible platform for autonomy education and research,'' in \emph{2017
  IEEE International Conference on Robotics and Automation (ICRA)}.\hskip 1em
  plus 0.5em minus 0.4em\relax IEEE, 2017, pp. 1497--1504.

\end{thebibliography}
%


\end{document}